\documentclass{article}

\PassOptionsToPackage{numbers, compress}{natbib}

\usepackage[preprint]{neurips_2021}

\usepackage[utf8]{inputenc} 
\usepackage[T1]{fontenc}    
\usepackage[pagebackref=true,breaklinks=true,colorlinks,bookmarks=false]{hyperref}       
\usepackage{url}            
\usepackage{booktabs}       
\usepackage{amsfonts}       
\usepackage{nicefrac}       
\usepackage{microtype}      
\usepackage{xcolor}         
\usepackage{graphicx}
\usepackage{multirow}
\usepackage{amsmath}

\title{Shuffle Transformer: Rethinking Spatial Shuffle for Vision Transformer}

\author{%
    Zilong Huang,  Youcheng Ben,  Guozhong Luo, Pei Cheng, Gang Yu, Bin Fu \\
    Tencent GY-Lab \\
    \tt\small\{zilonghuang, eugeneben, alexantaluo, peicheng, skicyyu, brianfu\}@tencent.com \\
}

\begin{document}

\maketitle

\begin{abstract}
  Very recently, Window-based Transformers, which computed self-attention within non-overlapping local windows, demonstrated promising results on image classification, semantic segmentation, and object detection. However, less study has been devoted to the cross-window connection which is the key element to improve the representation ability. In this work, we revisit the spatial shuffle as an efficient way to build connections among windows. As a result, we propose a new vision transformer, named Shuffle Transformer, which is highly efficient and easy to implement by modifying two lines of code. Furthermore, the depth-wise convolution is introduced to complement the spatial shuffle for enhancing neighbor-window connections. The proposed architectures achieve excellent performance on a wide range of visual tasks including image-level classification, object detection, and semantic segmentation. Code will be released for reproduction.
\end{abstract}

\section{Introduction}
Recently, Vision Transformers~\cite{dosovitskiy2020image,touvron2020deit,carion2020end,zheng2020rethinking} have been absorbed more and more investigations. Due to the great flexibility in modeling long-range dependencies in vision tasks, introducing less inductive bias, vision transformers have already achieved performances quite competitive with their CNN counterparts~\cite{he2016deep,tan2019efficientnet}
and can naturally process multi-modality input data including images, videos, texts, speech signals,
and point clouds.

These vision transformers have quadratic computation complexity to input image size due to computation of self-attention globally. This property makes it difficult to apply vision transformers to dense prediction tasks, such as semantic segmentation, which inherently require high-resolution images. A workaround is the window-based self-attention, where the input is spatially partitioned into non-overlapped windows and the standard self-attention is computed only within each local window. On the one hand, it can significantly reduce the complexity, On the other hand, limiting the range to compute self-attention can make
the model more efficient to train~\cite{beltagy2020longformer,zhu2020deformable}. However, computing self-attention within non-overlapping windows results in a limited receptive field. 
To build information communication across windows, HaloNet~\cite{vaswani2021scaling} scales the window for collecting information inside and outside the original window. Swin~\cite{liu2021Swin} alleviates this issue by shifting the window partition between consecutive self-attention layers. Despite being effective, HaloNet and Swin both are not good at building long-range cross-window connections for enlarging receptive fields efficiently. 

To build the cross-window connections, especially long-range cross-window connections, we pay attention to window partitioning. Inspired by ShuffleNet~\cite{zhang2018shufflenet}, which comes up with a novel channel shuffle operation to help the information flowing across feature channels, we introduce a spatial shuffle operation into the window-based self-attention module for providing connections among windows and significantly enhance modeling power. Different from channel shuffle, spatial shuffle operation will result in spatial misalignment between features and image content. Thus, we need an inverse process of spatial shuffle operator to align the feature with image content, which is named as spatial alignment operator. In this paper, the window-based transformer block with a spatial shuffle operator and a spatial alignment operator, named as Shuffle Transformer Block, will be used as the basic component to build the Shuffle Transformer. 

Although spatial shuffle is efficient for building long-range cross-window connections, the “grid issue” may arise when image size is far greater than the window size. To enhance the neighbor-window connections, we introduce a depth-wise convolutional layer with a residual connection into the Shuffle Transformer Block. Finally, with the help of successive Shuffle Transformer Blocks, the proposed Shuffle Transformer could make information flow across all windows.

In summary, our proposed Shuffle Transformer achieves linear computational complexity in the number of input tokens by computing self-attention within non-overlapping local windows. To build rich cross-window connections, we propose the Shuffle Transformer block which integrates the spatial shuffle and neighbor-window connections. The experiments conducted on a number of visual tasks, ranging from image-level
classification to pixel-level semantic/instance segmentation and object detection show that both of our proposed architectures perform favorably against other state-of-the-art vision transformers with similar computational complexity.

\section{Related Work}

\noindent\textbf{Vision Transformers}
The transformer was firstly proposed by~\cite{vaswani2017attention} for machine translation tasks and has dominated in natural language modeling. Recently, ViT~\cite{dosovitskiy2020image} is the first work to apply a pure transformer to image classification with state-of-the-art performance (e.g. ResNets~\cite{he2016deep}, EfficientNet~\cite{tan2019efficientnet}) on image classification when the data is large enough. It splits each image into a sequence of tokens and then applies multiple standard Transformer layers to model their global relation for classification. Specifically, the standard Transformer layers consist of a Multi-head Self-Attention module (MSA) and a Multiple Layer Perceptron (MLP). Following that, DeiT~\cite{touvron2020deit} introduces token-based distillation to reduce the data necessary for training the transformer. T2T-ViT~\cite{yuan2021tokens} structures the image to tokens by recursively aggregating neighboring tokens into one token to reduce tokens length. Transformer-iN-Transformer (TNT)~\cite{han2021transformer} utilizes both an outer Transformer block that processes the patch embeddings, and an inner Transformer block that models the relation among pixel embeddings, to model both patch-level and pixel-level representation.

To produce a hierarchical representation that is required by dense prediction tasks such as object detection and segmentation, Pyramid Vision Transformer (PVT)~\cite{wang2021pyramid}, is proposed and can output the feature pyramid~\cite{lin2017feature} as in convolutional neural networks (CNNs). 

To reduce the complexity, The recent Swin Transformer~\cite{liu2021Swin} introduces non-overlapping window partitions and restricts self-attention within each local window, resulting in linear computational complexity in the number of input tokens.

\noindent\textbf{Integrating Convolution and Vision Transformer}
To enhance local context in Vision Transformers, the Conditional Position encodings Visual Transformer (CPVT)~\cite{chu2021conditional} replaces the predefined positional embedding used in ViT with conditional
position encodings (CPE), enabling Transformers to process input images of arbitrary size without interpolation. CvT~\cite{wu2021cvt} and CCT~\cite{hassani2021escaping} introducing convolutions into the Vision Transformer architecture to merge the benefits of Transformers with the benefits of CNNs for image recognition tasks. LocalViT~\cite{li2021localvit} brings a locality mechanism to vision transformers by inserting introducing depth-wise convolutions into the MLP module. We also introduce convolution into the window-based vision transformer for enhancing the neighbor-window connections. Different from the others, we place the depth-wise convolution between the window-based multi-head self-attention and the MLP module.

\noindent\textbf{Window-based Self-Attention}
The standard Transformer architecture~\cite{vaswani2017attention} and its adaptation for image classification~\cite{dosovitskiy2020image} both conduct global self-attention, where the relationships for each token-pair are computed. The computation for dense relationships leads to quadratic complexity with respect to the number of tokens, making it unsuitable for the dense prediction tasks, such as
semantic segmentation, inherently requires high-resolution images.

For efficient modeling, there are some works apply local/window constraints to self-attention, proposed by ~\cite{huang2020ccnetv2,hu2019local,ramachandran2019stand,wang2020axial,child2019generating,liu2021Swin,huang2019interlaced, chu2021twins}, reduces the computation cost. These window-based self-attention based methods can be grouped into two categories: sliding-window based methods~\cite{huang2020ccnetv2,hu2019local,ramachandran2019stand,wang2020axial,child2019generating} and non-overlapping window-based methods~\cite{liu2021Swin,huang2019interlaced}. As mentioned in Swin~\cite{liu2021Swin}, compared with non-overlapping window-based methods, sliding window-based self-attention approaches suffer from low latency on general hardware due to different key sets for different query pixels.
However, non-overlapping window-based methods compute self-attention in the non-overlapping local window, which lacks cross-window connections. Swin~\cite{liu2021Swin} alleviates this issue by shifting the window partition between consecutive self-attention layers. The proposed Shuffle Transformer also utilizes non-overlapping window-based self-attention and takes spatial shuffle to make information flow across windows.

\noindent\textbf{Channel Shuffle and Spatial Shuffle}
Modern efficient convolutional neural networks~\cite{zhang2018shufflenet,han2020ghostnet,sandler2018mobilenetv2} apply group or depth-wise convolutions to reduce the complexity, which meets the issue of channel sparse connections. To exchange information of channels from different groups, ShuffleNet shuffles the channel to make cross-group information flow for multiple group convolution layers.

Following that, Spatially Shuffled Convolution~\cite{kishida2019incorporating} incorporate the random spatial shuffle in the regular convolution. The most related work to ours is Interlaced Sparse Self-Attention (ISSA)~\cite{huang2019interlaced}, which apply the interlacing mechanism to decompose the dense affinity matrix within the self-attention mechanism with the product of two sparse affinity matrices. However, the number of windows is fixed in ISSA, which also leads to quadratic complexity with respect to the input size. We fix the window size rather than the number of windows, which results in linear computational complexity. Besides, the motivations are also different. We adapt spatial shuffle for window-based vision transformers to bridge the connection among the non-overlapping windows. ISSA uses two sparse self-attention to mimic the global self-attention for semantic segmentation.

\begin{figure*}[!t]
    \centering
    \includegraphics[width=1.0\linewidth]{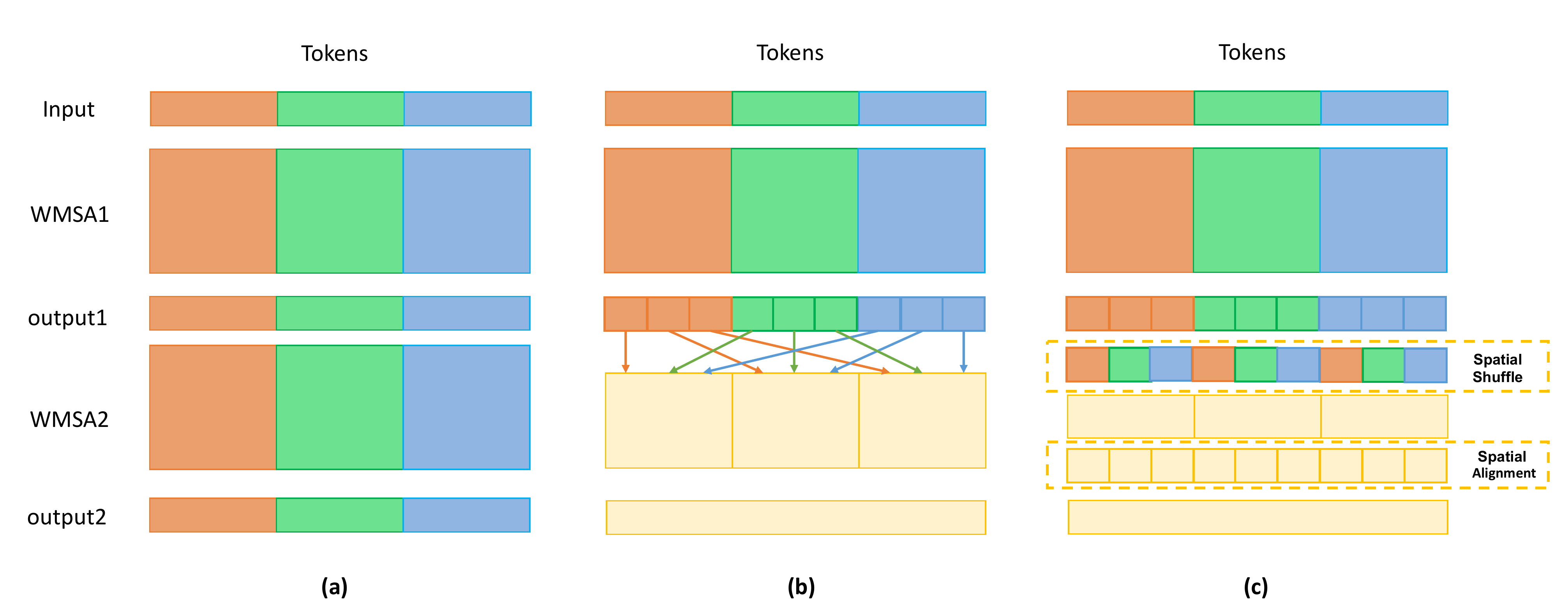}
    \caption{Spatial shuffle with two stacked window-based Transformer block. The MLP is omitted in the visualization because it does not affect the information interaction in the spatial dimension. WMSA stands for window-based multi-head self-attention. a) two stacked window-based Transformer blocks with the same window size. Each output token only relates to the tokens within the window. No cross-talk; b) tokens from different windows are fully related when WMSA2 takes data from different windows after WMSA1; c) an equivalent implementation to b) using spatial shuffle and alignment. 
    }
    \label{fig:spatial_shuffle}
    \vspace{0mm}
\end{figure*}

\section{Shuffle Transformer}
In this section, we start from the standard window-based multi-head self-attention. To build long-range cross-window talks, the spatial shuffle is proposed. The neighbor-window connection module is used to enhance the connections of neighborhood windows. Then, we integrate the spatial shuffle and neighbor-window connection module into the shuffle transformer block for building rich cross-window connections. Finally, the overall network architecture with its variants is given.

\subsection{Window-based Multi-head Self-Attention}
For efficient modeling, there are some works that apply local constraints to self-attention, proposed by~\cite{hu2019local,ramachandran2019stand,wang2020axial,liu2021Swin}, reduces the computation cost. Window-based Multi-head Self-Attention (WMSA) is proposed to compute self-attention within local windows. The windows are arranged to evenly partition the image in a non-overlapping manner. Supposing each window contains $M \times M$ tokens, the computational complexity of a global MSA module and a window-based one on an input of $H \times W$ tokens with dimension $C$ are $\mathcal{O}(H^2W^2C)$ and $\mathcal{O}(M^2HWC)$, respectively. Thus, the Window-based Multi-head Self-Attention is significantly more efficient when $M \ll H$ and $M \ll W$ and grows linearly with $HW$ if $M$ is fixed. Compared with a global MSA module, the WMSA module needs a window partition operation before computing self-attention and a windows-to-image operation after computing self-attention. However, the computation cost of the additional operations is negligible in practical implementation. Except for the window partition, WMSA shares the same structure with the global MSA module.

\subsection{Spatial Shuffle for Cross-Window Connections} \label{Spatial_Shuffle}

Although the window-based multi-head self-attention is computation friendly. However, the image is divided into non-overlapping sub-windows. If multiple window-based self-attention modules stack together, there is one side effect: the receptive field is limited within the window, which has an adverse effect on tasks such as segmentation, especially on high-resolution inputs. Fig~\ref{fig:spatial_shuffle}(a) illustrates a situation of two stacked window-based self-attention. It is clear that outputs from a certain window only relate to the inputs within the window. This property blocks information flow among windows and weakens representation.

To address the issue, a straightforward solution is to allow the second window-based self-attention to obtain input data from different windows (as shown in Fig~\ref{fig:spatial_shuffle}(b)), the tokens in different windows will be related. Inspired by ShuffleNet~\cite{zhang2018shufflenet}, this can be efficiently and elegantly implemented by a spatial shuffle operation (Fig~\ref{fig:spatial_shuffle} (c)). Without loss of generality, input is assumed as the 1D sequence. Suppose a Window-based Self-Attention with window size $M$ whose input has $N$ tokens; we first reshape the output spatial dimension into ($M, \frac{N}{M}$), transposing and then flattening it back as the input of the next layer. 
This kind of operation puts the tokens from distant windows together and helps to build long-range cross-window connections. Different from channel shuffle, spatial shuffle needs the spatial alignment operation to adjust the spatial tokens into the original positions for aligning features and image content spatially. The spatial alignment operation first reshapes the output spatial dimension into ($\frac{N}{M}, M$), transposing and then flattening it, which is an inverse process of the spatial shuffle. 

Considering there are always the window partition operation and windows-to-image operation before and after computing self-attention, we could merge the window partition operation with the spatial shuffle operation, windows-to-image operation with the spatial alignment operation, thus the spatial shuffle operation and spatial alignment do not bring extra computation and are easy to implement by modifying two lines of code.  Moreover, spatial shuffle is also differentiable, which means it can be embedded into network structures for end-to-end training. 

\begin{figure*}[!t]
    \centering
    \includegraphics[width=1.0\linewidth]{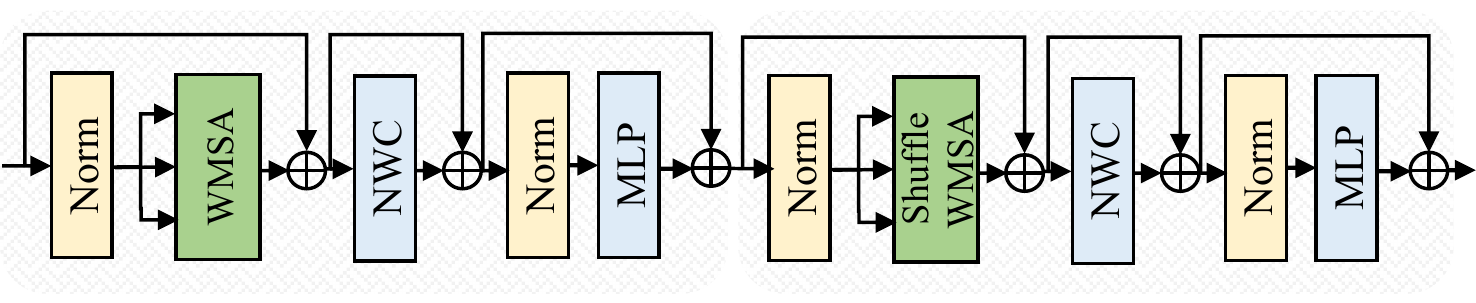}
    \caption{Two successive Shuffle Transformer Block. The WMSA and Shuffle WMSA are window-based multi-head self attention without/with spatial shuffle, respectively.
    }
    \label{fig:shuffle_transformer_block}
    \vspace{0mm}
\end{figure*}

\subsection{Neighbor-Window Connection Enhancement}
Bringing the spatial shuffle into the window-based multi-head self-attention could build cross-window connections, especially long-range cross-window. However, there is an underlying apprehension when processing a high-resolution image. The ``grid issue'' may arise when image size is far greater than the window size. 

Fortunately, there are several approaches to this issue by enhancing the neighbor-window connections. 1) scale the window size~\cite{vaswani2021scaling}. 2) cooperate with shifted window~\cite{liu2021Swin}. 3) introduce convolution to shuffle transformer block~\cite{li2021localvit,wu2021cvt}. Considering the efficiency, we insert a depth-wise convolution layer with a residual connection between the WMSA module and the MLP module. The kernel size of the depth-wise convolution layer is the same as the window size. This operator could strengthen the information flow among nearby windows and alleviate the ``grid issue''. To find a better place for a depth-wise convolution layer, we conduct the ablation studies, as shown in Table~\ref{fig:my_label}.

\subsection{Shuffle Transformer Block}
The Shuffle Transformer Block consists of the Shuffle Multi-Head Self-Attention module (Shuffle-MHSA), the Neighbor-Window Connection module (NWC), and the MLP module.
To introduce cross-window connections while maintaining the efficient computation of non-overlapping windows, we propose a strategy which alternates between WMSA and Shuffle-WMSA in consecutive Shuffle Transformer blocks. As shown in Figure~\ref{fig:shuffle_transformer_block}. The first window-based transformer block uses regular window partition strategy and the second window-based transformer block uses window-based self-attention with spatial shuffle. Besides, the Neighbor-Window Connection moduel (NWC) is added into each block for enhancing connections among neighborhood windows. Thus the proposed shuffle transformer block could build rich cross-window connections and augments representation.  Finally, the consecutive Shuffle Transformer blocks are computed as
\begin{align*} \label{eq:optimizing}
& x^l=\textbf{WMSA}(\textbf{BN}(z^{l-1}))+z^{l-1}, \nonumber\\
& y^l=\textbf{NWC}(x^{l})+x^{l}, \nonumber\\
& z^l=\textbf{MLP}(\textbf{BN}(y^{l}))+y^{l}, \nonumber\\
& x^{l+1}=\textbf{Shuffle-WMSA}(\textbf{BN}(z^l))+z^l, \nonumber\\
& y^{l+1}=\textbf{NWC}(x^{l+1})+x^{l+1}, \nonumber\\
& z^{l+1}=\textbf{MLP}(\textbf{BN}(y^{l+1}))+y^{l+1}.
\end{align*}

where $z^l$, $y^l$ and $z^l$ denote the output features of the (Shuffle-)WMSA module, the Neighbor-Window Connection module and the MLP module for block $l$, respectively; WMSA and Shuffle-WMSA denote window-based multi-head self-attention without/with spatial shuffle, respectively. To better handle 2D input, we adapt the standard transformer block by replacing the Layernorm~\cite{ba2016layer} with Batchnorm~\cite{ioffe2015batch}. Meanwhile, the Linear layer in Shuffle WMSA and MLP is changed to a convolutional layer with kernel size $1\times1$.

\begin{figure*}[!t]
    \centering
    \includegraphics[width=1.0\linewidth]{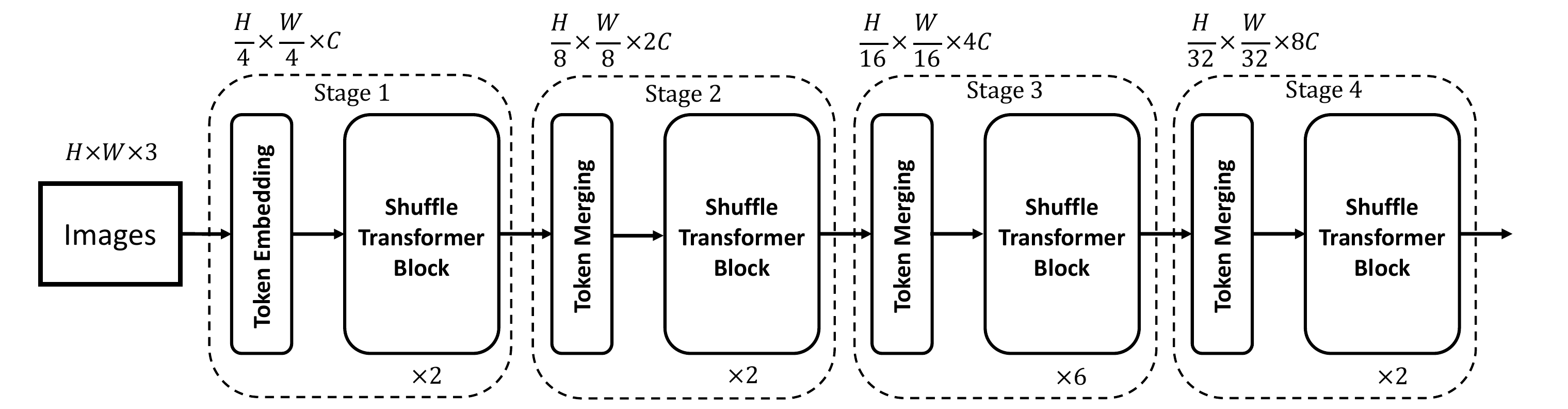}
    \caption{The architecture of a Shuffle Transformer (Shuffle-T).}
    \label{fig:shuffle_transformer}
    \vspace{0mm}
\end{figure*}

\subsection{Architecture and Variants}
An overview of the Shuffle Transformer architecture is presented in Figure~\ref{fig:shuffle_transformer}, which illustrates the tiny version (Shuffle-T). The Shuffle Transformer consists of a token embedding layer and several shuffle transformer blocks and token merging layers. In our implementation, we use two stacked convolution layers as token embedding layers. To produce a hierarchical representation, we use a convolutional layer with kernel size $2\times2$ and stride 2 as the token merging layer to reduce the number of tokens.

For a fair comparison, we follow the settings of swin~\cite{liu2021Swin}. We build our base model, called Shuffle-B, to have the model size and computation complexity similar to Swin-B/ViTB/DeiT-B. We also introduce Shuffle-T and Shuffle-S,
which are similar to Swin-T and Swin-S, respectively. The
window size is set to $M = 7$ by default. The query dimension of each head is d = 32, and the expansion layer of
each MLP is $\alpha = 4$, for all experiments. The architecture
hyper-parameters of these model variants are: Shuffle-T: $C = 96$, layer numbers = \{2, 2, 6, 2\}. Shuffle-S: $C = 96$, layer numbers = \{2, 2, 18, 2\}. Shuffle-B: $C = 128$, layer numbers = \{2, 2, 18, 2\}. 

\section{Experiments}

\begin{table}
\caption{ Comparison of different backbones on ImageNet-$1K$ classification. Throughput is measured with the batch size of $192$ on a single V100 GPU. All models are trained and evaluated on $224\times224$ resolution. }
\label{tab:cls}
\begin{center}
\scalebox{1.0}{
    \begin{tabular}{c c c c c}
    \toprule
    Method & Params & GFLOPs & Throughput (Images / s) & Top-1 (\%) \\ 
    \hline
    \multicolumn{5}{c}{ConvNet} \\
    \hline
    ReGNetY-4G~\cite{radosavovic2020designing} & 21M & 4.0 & 1157 & 80.0 \\
    ReGNetY-8G~\cite{radosavovic2020designing} & 39M & 8.0 & 592 & 81.7 \\
    ReGNetY-16G~\cite{radosavovic2020designing} & 84M & 16.0 & 335 & 82.9 \\
    \hline
    \multicolumn{5}{c}{Transformer} \\
    \hline
    DeiT-S/16~\cite{touvron2020deit} & 22M & 4.6 & 437 & 79.9 \\
    CrossViT-S~\cite{chen2021crossvit} & 27M & 5.6 & - & 81.0 \\
    T2T-ViT-14~\cite{yuan2021tokens} & 22M & 5.2 & - & 81.5 \\
    TNT-S~\cite{han2021transformer} & 24M & 5.2 & - & 81.3 \\
    PVT-Small~\cite{wang2021pyramid} & 25M & 3.8 & 820 & 79.8 \\
    Swin-T~\cite{liu2021Swin} & 29M & 4.5 & 766 & 81.3 \\
    Shuffle-T (\textbf{ours}) & 29M & 4.6 & 791 & $\mathbf{82.5}$ \\
    \hline
    T2T-ViT-19~\cite{yuan2021tokens} & 39M & 8.9 & - & 81.9 \\
    PVT-Medium~\cite{wang2021pyramid} & 44M & 6.7 & 526 & 81.2 \\
    Swin-S~\cite{liu2021Swin} & 50M & 8.7 & 444 & 83.0 \\
    Shuffle-S (\textbf{ours}) & 50M & 8.9 & 450 & $\mathbf{83.5}$ \\
    \hline
    ViT-B/16~\cite{dosovitskiy2020image} & 87M & 17.6 & 86 & 77.9 \\
    DeiT-B/16~\cite{touvron2020deit} & 87M & 17.6 & 292 & 81.8 \\
    T2T-ViT-24~\cite{yuan2021tokens} & 64M & 14.1 & - & 82.3 \\
    CrossViT-B~\cite{chen2021crossvit} & 105M & 21.2 & - & 82.2 \\
    TNT-B~\cite{han2021transformer} & 66M & 14.1 & - & 82.8 \\
    PVT-Large~\cite{wang2021pyramid} & 61M & 9.8 & 367 & 81.7 \\
    Swin-B~\cite{liu2021Swin} & 88M & 15.4 & 275 & 83.3 \\
    Shuffle-B (\textbf{ours}) & 88M & 15.6 & 279 & $\mathbf{84.0}$ \\
    \bottomrule
    \end{tabular}
}
\end{center}
\end{table}

To showcase the effectiveness of our approach, we conduct comprehensive experiments on three different tasks: ImageNet-$1K$ image classification~\citep{deng2009imagenet}, ADE$20K$ semantic segmentation~\cite{zhou2017scene} and COCO instance segmentation~\cite{lin2014microsoft}. 


\subsection{Classification on ImageNet-1K}
\textbf{Settings}~~For image classification, we evaluate the proposed Shuffle Transformer on ImageNet-$1K$~\cite{deng2009imagenet}, which contains $1.28M$ training images and $50K$ validation images from $1,000$ classes. We follow~\cite{liu2021Swin} to adopt regular $1K$ setting and report the top-$1$ accuracy on a single crop.
To be specific, we employ an AdamW~\cite{loshchilov2017decoupled} optimizer for $300$ epochs using a cosine decay learning rate scheduler and $20$ epochs of linear warm-up. A batch size of $1,024$, an initial learning rate of $0.001$, and a weight decay of $0.05$ are used. We also utilize the same augmentation and regularization strategies as shown in~\cite{liu2021Swin} for a fair comparison.


\textbf{Results}~~Table~\ref{tab:cls} presents detailed comparisons between our Shuffle Transformer and other backbones used for the training of ImageNet classification, including both ConvNet-based and Transformer-based approaches. All models are trained and evaluated on $224\times224$ resolution. Compared to the state-of-the-art ConvNets, i.e. RegNet~\cite{radosavovic2020designing}, our Shuffle Transformer achieves a better speed-accuracy trade-off. Meanwhile, when compared to existing Transformer-based approaches, e.g. Swin~\cite{liu2021Swin}, our Shuffle Transformer stably outperforms the counterparts with similar computational complexity.


\subsection{Semantic Segmentation on ADE20K}
\textbf{Settings}~~ADE$20K$~\cite{zhou2017scene} is a challenging scene parsing dataset for semantic segmentation. There are $150$ classes and diverse scenes with $1,038$ image level labels. The dataset is divided into three sets ($20,210$ / $2,000$ / $3,352$) for training, validation and testing. 

For fair comparison with the recent state-of-the-art model Swin~\cite{liu2021Swin}, we follow it to adopt UperNet~\cite{xiao2018unified} as the base framework for training. The AdamW~\cite{loshchilov2017decoupled} optimizer with an initial learning rate of $6\times10^{-5}$ and a weight decay of $0.01$ is used. We also utilize the warm-up during the first $1,500$ iterations. All models are trained for $160K$ iterations with a batch size of $16$. Data augmentation contains random horizontal flip, random resizing with a scale range of $[0.5, 2.0]$, and random cropping with a crop size of $512\times512$. For quantitative evaluation, the mean of class-wise intersection-over-union (mIoU) is used for accuracy comparison, and the number of float-point operations (FLOPs) and frames per second (FPS) are adopted for speed comparison. Results of both single-scale and multi-scale testing are reported with the scaling factor ranging from $0.5$ to $1.75$. 


\textbf{Results}~~As shown in Table~\ref{tab:seg}, we prepare different variants of the Shuffle Transformer for detailed comparison with Swin~\cite{liu2021Swin}. The results show that our three models (Shuffle-T/S/B) consistently achieve better mIoU performance than Swin with comparable inference speed. To be specific, under multi-scale testing, Shuffle-T outperforms Swin-T by $1.4\%$ mIoU, Shuffle-B achieves new state-of-the-art result $50.5\%$ mIoU which outperforms Swin-B by $0.8\%$ mIoU. Shuffle-S also achieves comparable performance to Swin-S.

\begin{table}
\caption{ Results of semantic segmentation on the ADE$20K$ validation set. $\dagger$ indicates that the model is pretrained on ImageNet-$22K$. FLOPs is measured on $1024\times1024$ resolution. $*$ indicates the FPS reproduced by us and is measured on $512\times512$ resolution. }
\label{tab:seg}

\begin{center}
\scalebox{1.0}{
\begin{tabular}{c c c c c c}
\toprule
Method & Backbone & Params & GFLOPs & FPS & mIoU / MS mIoU (\%) \\
\hline
DANet~\cite{fu2019dual} & ResNet-101 & 69M & 1119 & 15.2 & 43.6 / 45.2  \\
CCNet~\cite{huang2019ccnet} & ResNet-101 & 64M & 981 & 15.0 & 44.3 / 45.7 \\
DeepLabv3+~\cite{chen2018encoder} & ResNet-101 & 63M & 1021 & 16.0 & 45.5 / 46.4 \\
AlignSeg~\cite{alignseg} & ResNet-101 & 67M & 956 & 20.0 & 44.7 / 46.0 \\
OCRNet~\cite{yuan2019object} & ResNet-101 & 56M & 923 & 19.3 & - / 45.3 \\
UperNet~\cite{xiao2018unified} & ResNet-101 & 86M & 1029 & 20.1 & 43.8 / 44.9 \\
\hline
OCRNet~\cite{yuan2019object} & HRNet-w48 & 71M & 664 & 12.5 & - / 45.7 \\
DeepLabv3+~\cite{chen2018encoder} & ResNeSt-101 & 66M & 1051 & 11.9 & - / 46.9 \\
SETR~\cite{zheng2020rethinking} & ViT-Large$^{\dagger}$ & 308M & - & - & 48.6 / 50.3 \\
\hline
\multirow{2}{*}{UperNet}
 & Swin-T~\cite{liu2021Swin} & 60M & 945 & 28.6$^*$ & 44.5 / 45.8 \\   
 & Shuffle-T~\textbf{(ours)} & 60M & 949 & 30.1$^*$ & $\mathbf{46.6 / 47.6}$ \\
\hline
\multirow{2}{*}{UperNet}
 & Swin-S~\cite{liu2021Swin} & 81M & 1038 & 21.9$^*$ & 47.6 / 49.5 \\   
 & Shuffle-S~\textbf{(ours)} & 81M & 1044 & 22.6$^*$ & $\mathbf{48.4 / 49.6}$ \\
\hline
\multirow{2}{*}{UperNet}
 & Swin-B~\cite{liu2021Swin} & 121M & 1188 & 19.9$^*$ & 48.1 / 49.7 \\  
 & Shuffle-B~\textbf{(ours)} & 121M & 1196 & 21.4$^*$ & $\mathbf{49.0 / 50.5}$ \\
\bottomrule
\end{tabular}
}
\end{center}
\end{table}

\begin{table}
\caption{ Object detection and instance segmentation performance on the COCO val2017 dataset using the Mask R-CNN and Cascade Mask R-CNN framework. FLOPs is evaluated on $1280\times800$ resolution. }
\label{tab:det}

\begin{center}
\scalebox{1.0}{
\begin{tabular}{c | c c c | c c c | c c}
\toprule
Backbone & AP$^{b}$ & AP$^{b}_{50}$ & AP$^{b}_{75}$ & AP$^{m}$ & AP$^{m}_{50}$ & AP$^{m}_{75}$ & Params & GFLOPs \\ 
\hline
\multicolumn{9}{c}{Mask R-CNN} \\
\hline
ResNet50~\cite{he2016deep} & 41.0 & 61.7 & 44.9 & 37.1 & 58.4 & 40.1 & 44M & 260 \\
PVT-Small~\cite{wang2021pyramid} & 43.0 & 65.3 & 46.9 & 39.9 & 62.5 & 42.8 & 44M & 245 \\
Swin-T~\cite{liu2021Swin} & 46.0 & 68.2 & 50.2 & 41.6 & 65.1 & 44.8 & 48M & 264 \\
Shuffle-T\textbf{(ours)} & $\mathbf{46.8}$ & $\mathbf{68.9}$ & $\mathbf{51.5}$ & $\mathbf{42.3}$ & $\mathbf{66.0}$ & $\mathbf{45.6}$ & 48M & 268 \\
\hline
ResNet101~\cite{he2016deep} & 42.8 & 63.2 & 47.1 & 38.5 & 60.1 & 41.3 & 63M & 336 \\
PVT-Medium~\cite{wang2021pyramid} & 44.2 & 66.0 & 48.2 & 40.5 & 63.1 & 43.5 & 64M & 302 \\
Swin-S~\cite{liu2021Swin} & $\mathbf{48.5}$ & $\mathbf{70.2}$ & $\mathbf{53.5}$ & 43.3 & 67.3 & 46.6 & 69M & 354 \\
Shuffle-S\textbf{(ours)} & 48.4 & 70.1 & 53.5 & $\mathbf{43.3}$ & $\mathbf{67.3}$ & $\mathbf{46.7}$ & 69M & 359 \\
\hline
\multicolumn{9}{c}{Cascade Mask R-CNN} \\
\hline
DeiT-S~\cite{touvron2020deit} & 48.0 & 67.2 & 51.7 & 41.4 & 64.2 & 44.3 & 80M & 889 \\  
ResNet50~\cite{he2016deep} & 46.3 & 64.3 & 50.5 & 40.1 & 61.7 & 43.4 & 82M & 739 \\  
Swin-T~\cite{liu2021Swin} & 50.5 & 69.3 & 54.9 & 43.7 & 66.6 & 47.1 & 86M & 745 \\  
Shuffle-T\textbf{(ours)} & $\mathbf{50.8}$ & $\mathbf{69.6}$ & $\mathbf{55.1}$ & $\mathbf{44.1}$ & $\mathbf{66.9}$ & $\mathbf{48.0}$ & 86M & 746 \\
\hline
ResNext101-32~\cite{xie2017aggregated} & 48.1 & 66.5 & 52.4 & 41.6 & 63.9 & 45.2 & 101M & 819 \\  
Swin-S~\cite{liu2021Swin} & 51.8 & 70.4 & 56.3 & 44.7 & $\mathbf{67.9}$ & 48.5 & 107M & 838 \\  
Shuffle-S\textbf{(ours)} & $\mathbf{51.9}$ & $\mathbf{70.9}$ & $\mathbf{56.4}$ & $\mathbf{44.9}$ & 67.8 & $\mathbf{48.6}$ & 107M & 844 \\
\hline
ResNext101-64~\cite{xie2017aggregated} & 48.3 & 66.4 & 52.3 & 41.7 & 64.0 & 45.1 & 140M & 972 \\  
Swin-B~\cite{liu2021Swin} & 51.9 & 70.9 & 56.5 & 45.0 & 68.4 & 48.7 & 145M & 982 \\  
Shuffle-B\textbf{(ours)} & $\mathbf{52.2}$ & $\mathbf{71.3}$ & $\mathbf{57.0}$ & $\mathbf{45.3}$ & $\mathbf{68.5}$ & $\mathbf{48.9}$ & 145M & 989 \\
\bottomrule
\end{tabular}
}
\end{center}
\vspace{-0mm}
\end{table}

\subsection{Instance Segmentation on COCO}
\textbf{Settings}~~Experiments of object detection and instance segmentation are conducted on COCO 2017 dataset~\cite{lin2014microsoft}, which contains $118K$ training, $5K$ validation and $41K$ test images. We follow~\cite{liu2021Swin} to evaluate the performance of our method based on Mask R-CNN and Cascade Mask R-CNN~\cite{he2017mask,cai2018cascade}. To be specific, we replace the backbones of these detectors with our shuffle transformer blocks. All the models are trained under the same setting as in~\cite{liu2021Swin}: multi-scale training~\cite{carion2020end,sun2020sparse} (resizing the input such that the shorter side is between $480$ and $800$ while the longer side is at most $1,333$), AdamW~\cite{loshchilov2017decoupled} optimizer (initial learning rate of $0.0001$, weight decay of $0.05$, and batch size of $16$), and $3\times$ schedule ($36$ epochs).

\textbf{Results}~~As shown in Table~\ref{tab:det}, we compare our model to standard ConvNets, e.g. ResNe(X)t~\cite{xie2017aggregated}, as well as existing Transformer networks, e.g. Swin~\cite{liu2021Swin}, DeiT~\cite{touvron2020deit} and PVT-Medium~\cite{wang2021pyramid}. For comparison with the Mask R-CNN framework, Shuffle-T and Shuffle-S surpass ConvNets based methods and DeiT~\cite{touvron2020deit} and PVT-Medium~\cite{wang2021pyramid} by a large margin with comparable parameters size and GFLOPs. Besides, Compared with Swin Transformer, Shuffle Transformer achieve comparable results on all metrics. For comparison with the Cascade Mask R-CNN framework, Shuffle Transformer could stably outperform the other networks in the aspects of $AP^{b}$ and $AP^{m}$. The results indicate the effectiveness of the proposed Shuffle Transformer on the tasks of object detection and instance segmentation. 

\begin{table}
\caption{ Ablation study on the effect of spatial shuffle and the neighbor-window connection on two benchmarks, FLOPs is measured on $224\times224$ resolution. }
\label{tab:ab_sps}
\begin{center}
\scalebox{1.0}{
\begin{tabular}{c c c c c}
\toprule
Method & Params & GFLOPs & ImageNet Top-1 (\%) & ADE$20K$ mIoU (\%) \\ 
\hline
w/o shuffle & 28M & 4.5 & 80.2 & 41.7 \\
w/ shuffle & 28M & 4.5 & 81.5 & 43.8 \\
w/ shuffle \& NWC & 29M & 4.6 & 82.5 & 46.6 \\

\bottomrule
\end{tabular}
}
\end{center}
\vspace{-0mm}
\end{table}

\begin{table}
\caption{ Ablation study on different ways to spatial shuffle on two benchmarks. }
\label{tab:ab_long_range_shuffle}
\begin{center}
\scalebox{1.0}{
\begin{tabular}{c c c c c}
\toprule
Method & ImageNet Top-1 (\%) & ADE$20K$ mIoU (\%) \\ 
\hline
long-range spatial shuffle & 82.5 & 46.6 \\
short-range spatial shuffle & 81.9 & 45.2 \\
random spatial shuffle & 82.0 & 45.7 \\
\bottomrule
\end{tabular}
}
\end{center}
\vspace{-3mm}
\end{table}

\subsection{Ablation Studies}
We also perform comprehensive studies on the effectiveness of different design modules in our proposed Shuffle Transformer (Shuffle-T), using ImageNet-$1K$ image classification and UperNet on ADE$20K$ semantic segmentation. To be specific, we take the following aspects into consideration, i.e., the design of spatial shuffle operation, and the usage of the neighbor-window connection at different positions of the transformer block.

\begin{figure}
\begin{minipage}[b]{0.49\textwidth}
\centering
\includegraphics[width=1.0\linewidth]{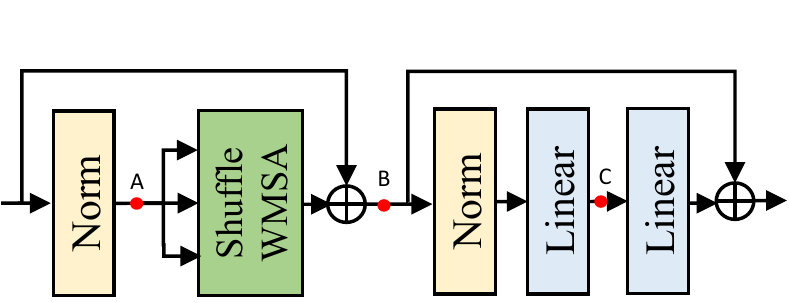}
\end{minipage}
\begin{minipage}[b]{0.49\textwidth}
\centering
\scalebox{0.77}{
\begin{tabular}[b]{c c c c c}
\toprule
\multirow{2}{*}{Position} & \multirow{2}{*}{Params} & \multirow{2}{*}{GFLOPs} & ImageNet & ADE$20K$ \\
& & & Top-1 (\%) &  mIoU (\%) \\ 
\hline
\textit{w/o NWC} & 28.3M & 4.5 & 81.5 & 43.8 \\
A & 28.5M & 4.6 & 81.9 & 46.2 \\
B & 28.5M & 4.6 & 82.5 & 46.6 \\
C & 29.2M & 4.8 & 82.3 & 46.3 \\
\bottomrule
\end{tabular}
}
\end{minipage}
\caption{Left: Visualization of three different positions to insert the neighbor-window connection. \textbf{A}: before the shuffle WMSA; \textbf{B}: after the residual connection of the shuffle WMSA; \textbf{C}: inside the MLP block. Right: Ablation study on the effect of the neighbor-window connection inserted at different positions, where A, B and C refer to three positions depicted left, and \textit{w/o NWC} means no neighbor-window connection is inserted. FLOPs is measured on $224\times224$ resolution.}
\label{fig:my_label}
\vspace{-3mm}
\end{figure}

\textbf{The effect of the spatial shuffle \& neighbor-window connection}~~Here, we discuss the influence of the spatial shuffle operation and the neighbor-window connection. Ablations of the spatial shuffle operation and the neighbor-window connection on the two tasks are reported in Table~\ref{tab:ab_sps}. Shuffle-T
with the spatial shuffle outperforms the counterpart built on a regular window-based multi-head self-attention used at each stage by +1.3\% top-1 accuracy on ImageNet-1K, +2.1 mIoU on ADE20K. The results indicate the effectiveness of using the spatial shuffle to build connections among windows in the preceding layers. Besides, adding the neighbor-window connection can further bring +1.0 \% top-1 accuracy on ImageNet-1K, +2.8 mIoU on ADE20K. The results indicate the effectiveness of using the neighbor-window connection to enhance the connection among neighborhood windows.

\textbf{The effect of the way to spatial shuffle}~Here, we discuss the influence of the way to spatial shuffle. Three kinds of spatial shuffles will be discussed. 1) long-range spatial shuffle is introduced in subsection~\ref{Spatial_Shuffle}. 2) Short-range spatial shuffle reshapes the output spatial dimension into ($\frac{N}{2M}, M, 2$), transposing and then flattening it. 3) random spatial shuffle will randomly shuffle the spatial dimension. As shown in Table~\ref{tab:ab_long_range_shuffle}, the long-range spatial shuffle achieves best performance on both image classification and segmentation tasks, which demonstrates the effectiveness of long-range spatial shuffle. And surprisingly, the random spatial shuffle also can achieve comparable performance.

\textbf{The effect of the position to insert the neighbor-window connection}~~We discuss the influence of inserting the neighbor-window connection at different positions of the transformer block. As shown in Figure~\ref{fig:my_label}, there are three positions A, B, and C used to place the neighbor-window connection (NWC), a depth-wise convolution layer with a residual connection.  The results indicate that inserting the neighbor-window connection between the shuffle WMSA and the MLP bock achieves the best performance.

\section{Conclusion}
In this paper, we have presented a Shuffle Transformer for a number of visual tasks, ranging from image-level
classification to pixel-level semantic/instance segmentation and object detection. For efficient modeling, we use the window-based multi-head self-attention which computes self-attention within the non-overlapping windows. To build cross-window connections, we introduce spatial shuffle into window-based multi-head self-attention. Meanwhile, to enhance the neighbor-window connections, we introduce a depth-wise convolutional layer with a residual connection into the Shuffle Transformer Block. Finally, with the help of successive Shuffle Transformer Block, the proposed Shuffle Transformer could make information flow across all windows. Extensive experiments show that both of our proposed architectures perform favorably against other state-of-the-art vision transformers with similar computational complexity. The simplicity and strong performance suggest that our proposed architectures may serve as stronger backbones for many vision tasks.

{\small
\bibliographystyle{plain}
\bibliography{main}
}

\end{document}